\documentclass{llncs}
\usepackage{amsmath,graphicx,booktabs}
\usepackage{epstopdf,url}
\usepackage{makeidx,amssymb}  
\usepackage{caption}

\begin{document}

\title{
ScanNet: A Fast and Dense Scanning Framework for Metastatic Breast Cancer  Detection from Whole-Slide Images}
%
%
\author{Huangjing Lin\inst{1}, Hao Chen\inst{1,2}, Qi Dou\inst{1}, Liansheng Wang\inst{3}, Jing Qin\inst{4}, Pheng-Ann Heng\inst{1}}
\institute{
The Chinese University of Hong Kong
\and Imsight Medical Technology Inc, China
\and Xiamen University, China
\and The Hong Kong Polytechnic University, Hong Kong
}
%


\maketitle

\begin{abstract}
Lymph node metastasis is one of the most significant diagnostic indicators in breast cancer, which is traditionally observed under the microscope by pathologists.
In recent years, computerized histology diagnosis has become one of the most rapidly expanding fields in medical image computing, which alleviates pathologists' workload and reduces misdiagnosis rate.
However, automatic detection of lymph node metastases from whole slide images remains a challenging problem, due to the large-scale data with enormous resolutions and existence of hard mimics.
In this paper, we propose a novel framework by leveraging fully convolutional networks for efficient inference to meet the speed requirement for clinical practice, while reconstructing dense predictions under different offsets for ensuring accurate detection on both micro- and macro-metastases.
Incorporating with the strategies of asynchronous sample prefetching and hard negative mining, the network can be effectively trained.
Extensive experiments on the benchmark dataset of \emph{2016 Camelyon Grand Challenge} corroborated the efficacy of our method. 
Compared with the state-of-the-art methods, our method achieved superior performance with a faster speed on the tumor localization task and surpassed human performance on the WSI classification task.
\end{abstract}

\section{Introduction}
\label{sec:intro}

Breast cancer has been one of the leading cancer killers threatening women in the world.
As one of the most important diagnostic criteria of breast cancer, detecting the metastases, especially in the sentinel lymph nodes, is a routine procedure for cancer staging performed by pathologists.
According to the pathologic TNM breast cancer staging system~\cite{edge2010american}, positive metastasis would lead to a higher staging of the patient, and afterwards necessary treatments would be accordingly arranged.
However, as is widely known, the process of pathologic diagnosis is extremely time-consuming and laborious, which requires pathologists to fully focus themselves hour by hour on the samples under the microscope.
Moreover, there is a considerable lack of pathologists amid the sharply growing demands of diagnosis with the cancer morbidity increasing~\cite{humphreys2010world}.
To the end, automatic, accurate and efficient detection algorithms are highly demanded.

\begin{figure}[t]
	\centering
	\includegraphics[width=.95\linewidth]{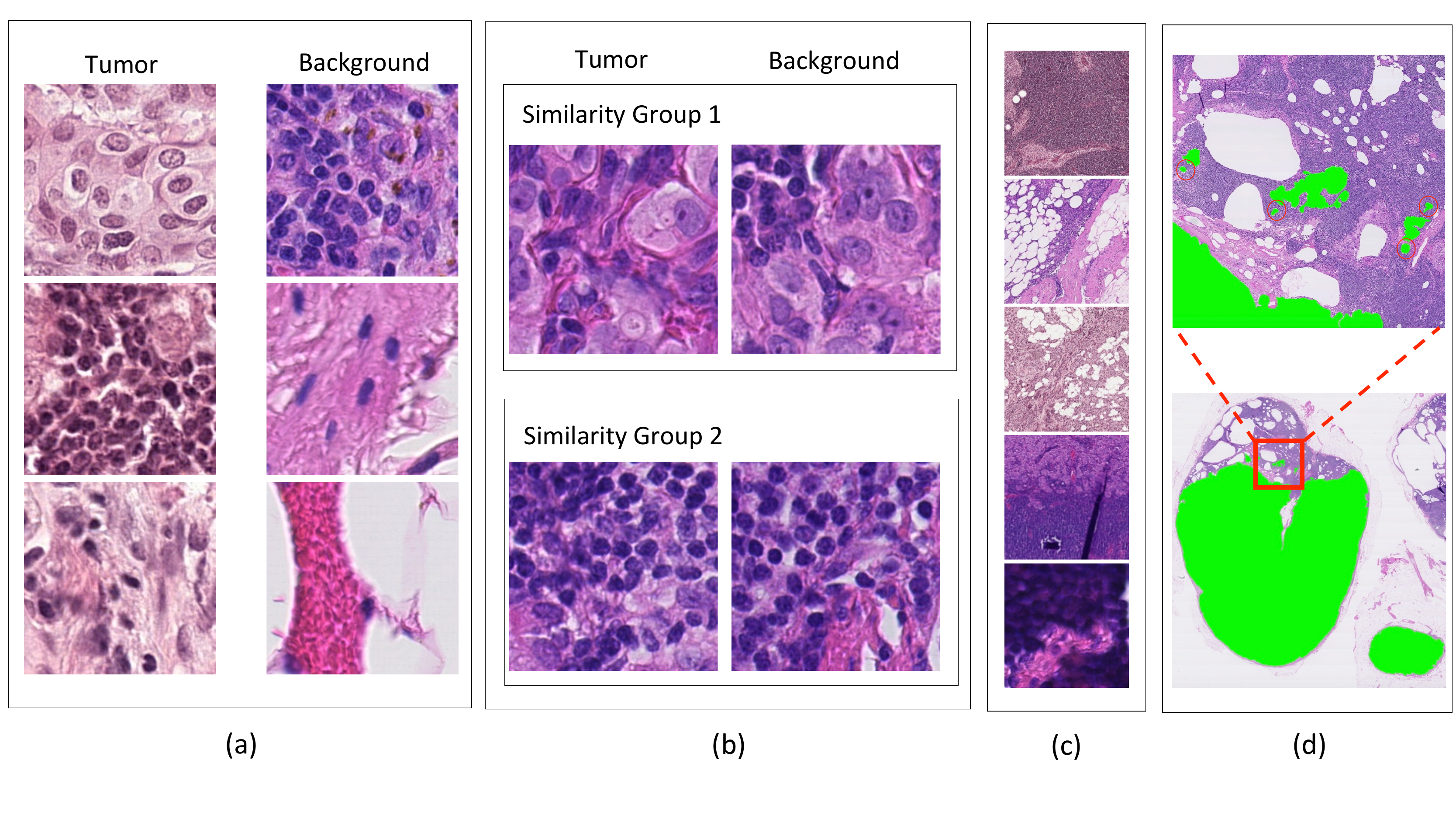}
	\vspace{-7mm}
	\caption{Illustration of the challenges for computerized histology image. (a) Variations of biological structures and textures. (b) Hard mimics from normal tissues in background. (c) Appearance variations due to the acquisition process. (d) Size difference between micro- and macro-metastases.}
	\label{fig:polypvarition}
\vspace{-8mm}
\end{figure}

Over the last decade, computerized histology analysis has been one of most rapidly expanding fields in medical image computing.
The computer aided diagnosis methods can not only alleviate pathologists' workload, but also contribute to reducing the misdiagnosis rate.
Nevertheless, automatic detection of metastases in lymph node digitized images is quite difficult with following challenges:
1) the large variations of biological structures and textures in both metastatic regions and background (Fig. \ref{fig:polypvarition}(a));
2) the hard mimics from normal tissues which carry similar morphological appearance with metastatic regions, shown in Fig. \ref{fig:polypvarition}(b);
3) as illustrated in Fig. \ref{fig:polypvarition}(c), the large variations introduced during the image acquisition process, e.g., sampling, staining and digitalizing;
4) the significant size variance between micro- and macro-metastases (Fig.~\ref{fig:polypvarition}(d)).
Last but not least, the resolution of whole-slide images (WSIs) is extremely huge,  approximating to $200,000\!\times\!100,000$ pixels. How to efficiently process such a giga-pixel image further poses challenges for automatic detection methods.

Many previous works have been dedicated to histopathological image analysis and achieved promising outcomes in different applications, such as mitosis detection~\cite{cirecsan2013mitosis,chen2016mitosis}, nuclei detection~\cite{sirinukunwattana2016locality} and colon gland segmentation~\cite{chen2017dcan}.
However, these studies were limited to regions of interest (e.g., image size $500\times500$) pre-selected by pathologists from WSIs.
To our best knowledge, there were limited works processing histology images in the whole-slide level until the~\emph{Camelyon Grand Challenge (Camelyon16)}~\cite{camelyon16} held in conjunction with 2016 ISBI recently.
In this challenge, Wang \emph{et al.} employed an ensemble of two GoogLeNets and achieved the best performance on metastasis detection~\cite{wang2016deep}. 
However, their framework used patch-based classifications, which would significantly increase the computation cost at the finest resolution (i.e., level-0), and hence is suboptimal in real clinical practice.
Recently, Xu \emph{et al.} proposed a sparse kernel technique to accelerate the pixel-wise predictions~\cite{xu2016detecting}, which could alleviate the efficiency problem to a certain extent. However, pixel-wise predictions would consume much more time and are not necessary in our underlying application. 



To tackle the aforementioned challenges, we propose a novel framework, referred as \emph{ScanNet}, based on fully convolutional network (FCN) for efficient inference and dense reconstruction for accurate metastasis detection from WSIs.
Superior to previous methods, our ScanNet can successfully eliminate the redundant computations compared to patch-based methods while generating dense predictions for improving the detection accuracy.
Specifically, our network is a modified deep FCN with a large receptive field in order to achieve efficient inference and accurate detection.
We further harness two effective training strategies to improve the discrimination capability and the time performance of the \emph{ScanNet}.
During testing, we propose a new strategy to make dense predictions by reconstructing the probability maps under different offsets.
Extensive experiments have been conducted on the benchmark dataset of~\emph{Camelyon16}.
Our ScanNet outperformed the state-of-the-art methods on the tumor localization task with a much faster speed, and achieved super-human performance on the WSI classification task, which validated the efficacy of our proposed framework.

\begin{figure}[t]
	\centering
	\includegraphics[width=1.0\linewidth]{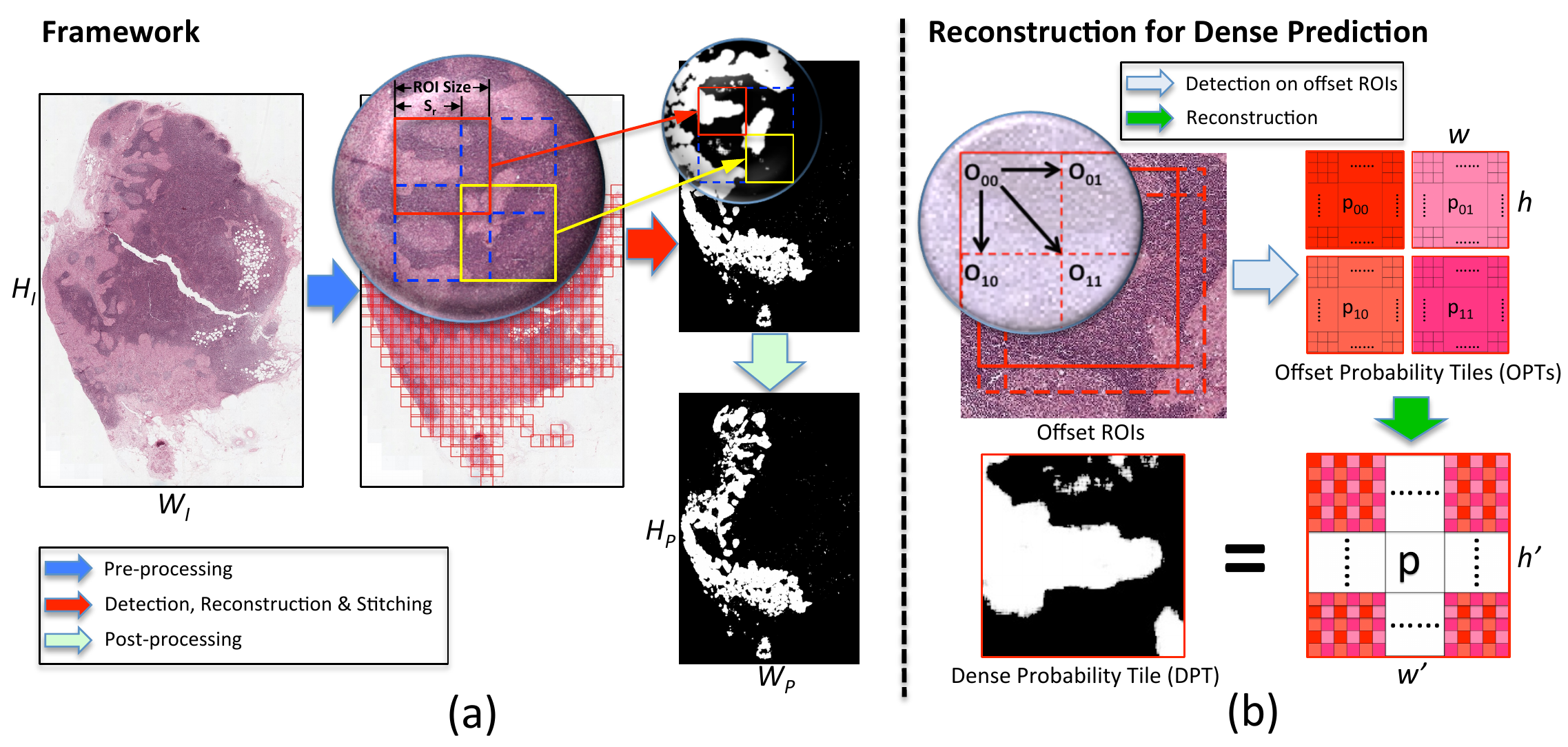}
	\caption{The illustration of our method. (a) The framework of our method. (b) Reconstruction for dense predictions. }
	\label{fig:method}
\vspace{-0.5cm}
\end{figure}
\section{Method}
\label{sec:method}
Fig.~\ref{fig:method}(a) illustrates the workflow of the proposed ScanNet.
We first employ a simple yet efficient method to remove the non-informative regions of input WSI.
Then we feed pre-processed images into the modified FCN equipped with reconstruction algorithm for efficient and dense predictions.
Finally, we utilize simple morphology operations to refine the results.


\subsection{Fast Metastasis Detection from WSIs via FCN}
\label{sec:method2-1}
\textbf{Pre-Processing.}
It is observed that more than $70\%$ area of a typical WSI is covered by the non-informative background that provides almost no information for cancer assessment.
In order to remove these regions to save computational cost, we employ the simple OTSU algorithm~\cite{otsu1975threshold} to determine the adaptive threshold and filter out most of the white background, as shown in Fig.~\ref{fig:method}(a).
In addition, to accelerate this operation, we conduct it using a multi-level mapping strategy.
That is we filter the downsampled (e.g., level-5) image first and then map the filtered image back into the original (level-0) image, which achieves dozen times of acceleration in the pre-processing step.
\\
\textbf{Fast Prediction via FCN}.
We propose to harness a modified fully convolutional network for fast prediction for large WSIs by taking its advantage of allowing us to take arbitrary sized images as input.
Different from traditional FCNs that are commonly used for segmentation tasks~\cite{chen2017dcan}, our FCN has no upsampling path which is a must for segmentation but not necessary for detection tasks.
In addition, the upsampling path would greatly slow down the detection process considering the large size of WSIs.
Using the FCN without upsampling path, our method can efficiently output a probability map, referred as probability tile, with a much smaller size than the input image.
We further leverage a reconstruction algorithm, which will be elaborated in the next subsection, to generate the dense probability map by stitching these tiles.
We implement the proposed FCN based on a modified VGG-16~\cite{simonyan2014very} network by removing padding operations and replacing the last three fully connected layers with fully convolutional layers $1024\times1024\times2$ (i.e., kernel size $1\times1$).
Thus, our FCN can enjoy the transferred features learned from a large set of natural images~\cite{chen2017dcan}. 
In the training process, we employ patch samples with size as $244\times244$ randomly cropped from WSIs to train the FCN.
Thanks to the merit of the fully convolutional mechanism, in the detection stage, we can take an image as large as $2868\times2868$ (determined by the capacity of GPU memory) as input and output a probability tile with size of $83\times83$.
By such a way, our ScanNet can process a WSI more than hundreds of times faster than traditional patch-based classification framework with the same stride.
\\
\textbf{Effective Training Strategies.}
We further propose two effective training strategies to enhance the learning process of FCN.
The first strategy is called \emph{Asynchronous Sample Prefetching}.
To save the memory space and augment the training samples flexibly, we generate the training samples on-the-fly in the data preparation process.
During the training phase, there exists a heavy I/O bottleneck, in which the GPU is often idle while waiting for the batched training data.
To resolve this problem, we implement this asynchronous sample prefetching mechanism by using multiple producer processes for CPU to prepare the training samples while one consumer process for GPU to consume the training data.
This strategy can keep GPU running all the time and boost at least $10$ times acceleration in the training stage.
The second strategy is called \emph{Hard Negative Mining}.
While there exist lots of negative training samples from the WSIs, most of them can be easily distinguished from the true metastases.
In order to enhance the discrimination capability of our ScanNet, we add the false positive samples, i.e., hard negative mining (HNM) examples, from the previously trained classifier back to the training data.
This strategy makes the training process more effectively by focusing on hard cases, which can help to significantly boost the recognition performance. 


\subsection{OPTs Reconstruction for Dense Prediction }
Reconstruction of the whole-slide probability map from the set of probability tiles generated by the FCN for accurate dense predictions is one of the key steps of the proposed ScanNet and it is not trivial.
As the set of probability tiles are generated from a set of sub-images extracted from the input WSI by a certain offset, we call the set of probability tiles as \emph{offset probability tiles} (OPTs).
We propose a two-stage scheme to reconstruct the whole-slide probability map.
We first generate a set of \emph{dense probability tiles} (DPTs) based on the OPTs and then stitch these DPTs together to obtain the final probability map.

The OPTs $p_{ij}$ are generated by the well trained FCN $\mathcal{F}$ given the input ROI image $I_r$ and the offsets of these images $\overrightarrow{O}_{ij}$, as illustrated in Fig.~\ref{fig:method}(b). 
We define the ratio between the size of DPTs and the size of OPTs as dense coefficient $\alpha$ and the OPTs can be formulated as:
\begin{equation}
\label {eq:OffsetDefinition}
\begin{cases}
 & p_{ij} = \mathcal{F}(I_r,\overrightarrow{O}_{ij}) \\
 & \overrightarrow{O}_{ij}=(i\ast S_{d},j\ast S_{d}), ~~~~~i,j\epsilon [0,1,...,\alpha-1] \\
\end{cases}
\end{equation}
where the $\overrightarrow{O}_{ij}$ is determined by the stride of DPTs $S_{d}$ and screen stride of FCN $S_{f}$, where $S_{d}=S_{f}/\alpha$ and $S_{f}=2^5$ (i.e., 5 pooling layers with a stride 2) in our FCN. 
Then we can calculate the DPTs by interweaving the OPTs alternatively.
Suppose that $(h', w')$ are the coordinates of a position in a DPT $p$, the probability of $p(h',w')$ can be calculated as:
\begin{equation}
\label {eq:DPMCalculation}
\begin{cases}
 & p(h',w')=p_{ij}(h,w) \\
 & i=h'~mod ~\alpha, ~~~j=w'~mod ~\alpha  \\
 & h=\left\lfloor h'/\alpha \right\rfloor, ~~~w= \left\lfloor w'/\alpha  \right\rfloor.        \\
\end{cases}
\end{equation}

After obtaining the set of DPTs, we stitch them together to generate the final probability map, as illustrated in Fig.~\ref{fig:method}(a).
To simplify the stitching process, the sub-images input to FCN should be extracted under a certain stride to ensure that all the DPTs are non-overlapped.
The stride of fetched sub-images, denoted as $S_{r}$, should satisfy the following constraint: $S_{r}=S_{d}\ast L_{p} $, where $L_{p}$ is the length size of the DPTs.
The positional relationship (take height $H$ for example and the width $W$ can be formulated in the same way) between the stitched probability map and the original WSI can be determined by following:
\begin{equation}
\label {eq:ProbMapping}
H_{I}=H_{P}\ast S_{d}+L_{i}/2
\end{equation}
where $H_{I}$ is the index of original WSI space, $H_{P}$ is the index of the stitched probability map, and $L_{i}$ is the input patch size in the training phase.

Finally, we post-processed the image by morphology opening operations to remove the small outliers.
For the localization task, each connected component in the binarized probability map (threshold was set as 0.5) was considered as a detection, with score equal to the maximum probability within the region.
For the WSI classification task, the prediction was simply computed as the maximal score within the slide without any sophisticated post-processing procedures.

\begin{table}[ht]
	\centering
\vspace{-0.5cm}
	\caption{Quantitative comparison with other methods}
	\label{tab:ComparisonResult}
\small
	\begin{tabular}{p{5cm} p{3cm} l c c c c c}
		\toprule
		Methods                  				&  FROC score & AUC score    \\
		\midrule
        Human performance            & - & 0.9660 \\
		\textbf{ScanNet-16(ours)}     &    \textbf{0.8533}    &    0.9875    \\
		HMS and MIT~\cite{wang2016deep}         &    0.8074  &    \textbf{0.9935}    \\
		HMS, Gordan Center, MGH       & 0.7600  & 0.9763   \\
		Radboud Uni. (DIAG)       &    0.5748  & 0.7786    \\
		EXB Research co.   	              &    0.5111  & 0.9156   \\
		Middle East Tech. Uni.   &   0.3889  & 0.8642 \\
		University of Toronto            & 0.3822 & 0.8149    \\
		DeepCare Inc                 & 0.2439  & 0.8833   \\
		NLP LOGIX co. USA             & 0.3859 & 0.8298 \\
		\midrule
		ScanNet-32(w/o HNM)   & 0.7030 &    0.9415   \\
		ScanNet-32  &  0.8133     &    0.9669      \\
		\bottomrule
	\end{tabular}
\\	
\vspace{-0.8cm}
\end{table}

\begin{figure}
	\centering
\vspace{-0.5cm}
	\includegraphics[width=1.\linewidth]{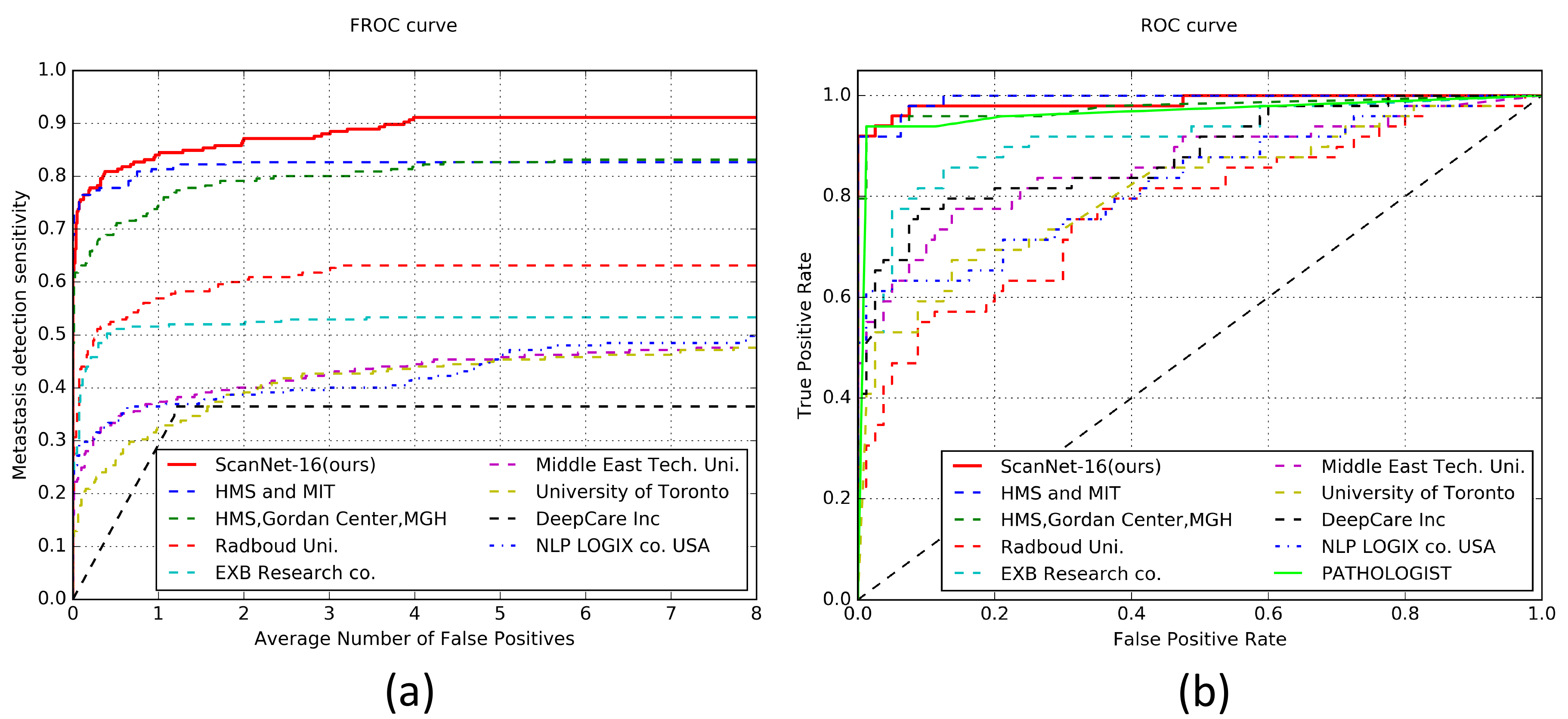}
	\caption{Evaluation Results. (a): FROC curves of tumor localization task of different methods. (b): ROC curves of WSI classification task of different methods.}
	\label{fig:TwoResults}
\vspace{-1cm}
\end{figure}

\section{Experiments and Results}
\label{sec:exp}
\begin{figure}[t]
	\centering
	\includegraphics[width=1.\linewidth]{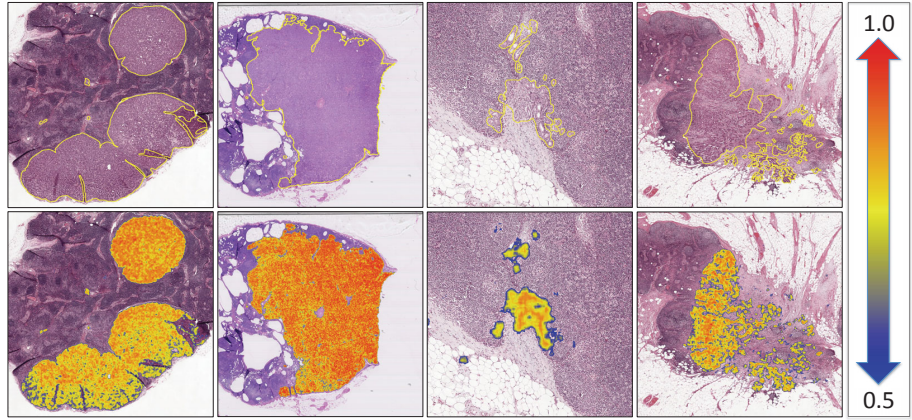}
	\caption{Typical examples of metastasis detection results from our proposed method. Top: ground truth annotations from pathologists indicated by the yellow lines. Bottom: Our detection results overlaid on the original images with different colors. } 
	\label{fig:QualitativeResults}
\vspace{-0.3cm}
\end{figure}

\textbf{Dataset.}
We evaluated our method on the benchmark dataset of \emph{Camelyon16} challenge~\cite{camelyon16}.
The training data contained $160$ normal WSIs and $110$ tumor WSIs with pixel-level annotations.
The testing data with $130$ WSIs were held out for independent evaluation.
To enrich the training samples, we augmented the training data with different transformations including translation, rotation, scaling, flipping and color jittering.
Our framework was implemented based on Caffe library~\cite{jia2014caffe}.
Typically, it took less than $15$ minutes to process one WSI with size $200,000\times100,000$ on a workstation with one Geforce GTX TITAN X GPU. 
\\
\textbf{Evaluation Metrics.}
The challenge consists of two tasks, i.e., tumor region localization and WSI classification.
The first task was evaluated based on Free Response Operating Characteristic (FROC) curve.
The FROC score was defined as the average sensitivity at 6 predefined false positive rates: 1/4, 1/2, 1, 2, 4 and 8 false positives per scan.
The second task was evaluated using the AUC score, i.e., area under the Response Operating Characteristic (ROC) curve.
\\
\textbf{Results and Comparison.}
In order to probe the efficacy of our method, we first evaluate our ScanNet under different configurations.
We set the $\alpha$ as $1$ and $2$ to produce the dense predictions in our ScanNet and call them ``ScanNet-32" and ``ScanNet-16", respectively.
Note that generally larger $\alpha$ means denser predictions can be generated from the network.
It is observed in Table~\ref{tab:ComparisonResult} that the results of ScanNet-16 is much better than that of ScanNet-32, demonstrating denser predictions can better detect the metastasis, especially micro cases.
We further evaluate the performance of ScanNet-32 with and without hard negative mining strategy and the results are listed in the last two rows of Table~\ref{tab:ComparisonResult}.
The performance of ScanNet-32 utilizing the hard negative mining strategy significantly outperforms its counterpart without the strategy on all metrics.
This demonstrates the efficacy of hard negative mining strategy to tackle the severe class imbalance problem with enormous negative samples which commonly happens in the field of medical image analysis.
%
We also compared our method with several state-of-the-art methods as shown in Table~\ref{tab:ComparisonResult}, and Fig.~\ref{fig:TwoResults} presents the FROC and ROC curves from different methods.
In the tumor localization task, our method achieved the best performance among all the methods, with the highest FROC score of 0.8533 outperforming the runner-up team~\cite{wang2016deep} by a significant margin of 4.6\%.
For the WSI classification task, our AUC score was 0.9875 with simple post-processing on tumor localization probability map, surpassing human performance of 0.9660 from the pathologists, and also quite competitive with the leading method of 0.9935~\cite{wang2016deep}.
Particular, we achieved the performance with a much faster speed than our competitors employing patch-based screening schemes.
We illustrate typical examples of metastasis detection results from our method in Fig.~\ref{fig:QualitativeResults}, from which we can observe the high consistency between our results and annotations from experienced pathologists.

\section{Conclusions}
In this paper, we proposed a novel fast and dense scanning framework for metastatic breast cancer detection from WSIs.
The proposed network can achieve fast processing of large-scale WSIs while generating dense predictions for ensuring detection accuracy.
Extensive experiments on the benchmark dataset corroborated the efficacy of our proposed method, outperforming all other methods on the tumor localization task and surpassing human performance on WSI classification task.
Our method is inherently general and can be easily extended to other biomarkers (e.g., mitosis) detection problems from WSIs.


\bibliographystyle{splncs}
\bibliography{polyprefs}
\end{document}